\documentclass{article}

\usepackage{microtype}
\usepackage{graphicx}
\usepackage{subfigure}
\usepackage{booktabs} 
\usepackage[colorinlistoftodos]{todonotes}

\usepackage{hyperref}
\hypersetup{
    urlcolor=red,
    }
\DeclareUrlCommand\url{\color{red}}

\usepackage{mathtools} 
\usepackage{mathcommand}
\usepackage{amssymb}
\usepackage{amsfonts} 

\providecommand\given{\vert}

\usepackage{algpseudocode}
\usepackage{algorithm}

\newcommand\MidSymbol[1][]{%
\nonscript\:#1
\allowbreak
\nonscript\:
\mathopen{}}

\DeclareMathOperator{\opInformationContent}{h}
\DeclarePairedDelimiterXPP{\ICof}[1]{\opInformationContent}{(}{)}{}{%
    \ifblank{#1}{\:\cdot\:}{#1}
}

\def\NoNumber#1{{\def\alglinenumber##1{}\State #1}\addtocounter{ALG@line}{-1}}

\DeclareMathOperator{\opEntropy}{H}
\DeclareMathOperator{\opPEntropy}{h}
\DeclarePairedDelimiterXPP{\Hof}[1]{\opEntropy}{[}{]}{}{%
    \renewcommand\given{\MidSymbol[\delimsize\vert]}
    \ifblank{#1}{\:\cdot\:}{#1}
}
\DeclarePairedDelimiterXPP{\hof}[1]{\opPEntropy}{[}{]}{}{%
    \renewcommand\given{\MidSymbol[\delimsize\vert]}
    \ifblank{#1}{\:\cdot\:}{#1}
}
\DeclarePairedDelimiterXPP{\xHof}[1]{\opEntropy}{(}{)}{}{%
    \ifblank{#1}{\:\cdot\:}{#1}
}

\DeclareMathOperator{\opMI}{I}
\DeclarePairedDelimiterXPP{\MIof}[1]{\opMI}{[}{]}{}{%
    \renewcommand\given{\MidSymbol[\delimsize\vert]}
    \ifblank{#1}{\:\cdot\:}{#1}
}
\DeclareMathOperator{\opPMI}{pmi}
\DeclarePairedDelimiterXPP{\PMIof}[1]{\opPMI}{[}{]}{}{%
    \renewcommand\given{\MidSymbol[\delimsize\vert]}
    \ifblank{#1}{\:\cdot\:}{#1}
}

\DeclarePairedDelimiterXPP{\CrossEntropy}[2]{\opEntropy}{(}{)}{}{%
    \ifblank{#1#2}{\:\cdot\: \MidSymbol[\delimsize\Vert] \:\cdot\:}{#1 \MidSymbol[\delimsize\Vert] #2}
}

\DeclareMathOperator{\opKale}{D_\mathrm{KL}}
\DeclarePairedDelimiterXPP{\Kale}[2]{\opKale}{(}{)}{}{%
    \ifblank{#1#2}{\:\cdot\: \MidSymbol[\delimsize\Vert] \:\cdot\:}{#1 \MidSymbol[\delimsize\Vert] #2}
}

\DeclareMathOperator{\opp}{p}
\DeclarePairedDelimiterXPP{\pof}[1]{\opp}{(}{)}{}{%
    \renewcommand\given{\MidSymbol[\delimsize\vert]}
    \ifblank{#1}{\:\cdot\:}{#1}
}

\DeclarePairedDelimiterXPP{\pcof}[2]{\opp_{#1}}{(}{)}{}{%
    \renewcommand\given{\MidSymbol[\delimsize\vert]}
    \ifblank{#2}{\:\cdot\:}{#2}
}

\DeclareMathOperator{\opq}{q}
\DeclarePairedDelimiterXPP{\qof}[1]{\opq}{(}{)}{}{%
    \renewcommand\given{\MidSymbol[\delimsize\vert]}
    \ifblank{#1}{\:\cdot\:}{#1}
}

\DeclarePairedDelimiterXPP{\varHof}[2]{\opEntropy_{\ifblank{#1}{\:\cdot\:}{#1}}}{[}{]}{}{%
    \renewcommand\given{\MidSymbol[\delimsize\vert]}
    \ifblank{#2}{\:\cdot\:}{#2}
}

\DeclarePairedDelimiterXPP{\xvarHof}[2]{\opEntropy_{\ifblank{#1}{\:\cdot\:}{#1}}}{(}{)}{}{%
    \renewcommand\given{\MidSymbol[\delimsize\vert]}
    \ifblank{#2}{\:\cdot\:}{#2}
}

\newcommand{\D}{\mathcal{D}}

\newcommand{\xval}{\textbf{x}^\text{val}}

\newcommand{\yval}{\textbf{y}^\text{val}}
\newcommand{\Yval}{\textbf{Y}^\text{val}}

\algdef{SE}[SUBALG]{Indent}{EndIndent}{}{\algorithmicend\ }%
\algtext*{Indent}
\algtext*{EndIndent}

\usepackage[accepted]{icml2021}

\icmltitlerunning{Learnable, Worth Learning, not yet Learned}

\begin{document}

\twocolumn[
\icmltitle{
Prioritized training on points that are learnable, \\worth learning, and not yet learned (workshop version)
}
            



\icmlsetsymbol{equal}{*}

\begin{icmlauthorlist}
\icmlauthor{S\"{o}ren Mindermann}{equal,ox}
\icmlauthor{Muhammed Razzak}{equal,ox}
\icmlauthor{Winnie Xu}{equal,to,co}
\icmlauthor{Andreas Kirsch}{ox}
\icmlauthor{Mrinank Sharma}{oxstats}
\icmlauthor{Adrien Morisot}{co}
\icmlauthor{Aidan N. Gomez}{ox,co}
\icmlauthor{Sebastian Farquhar}{ox}
\icmlauthor{Jan Brauner}{ox}
\icmlauthor{Yarin Gal}{ox}
\end{icmlauthorlist}

\icmlaffiliation{ox}{OATML, University of Oxford, United Kingdom}
\icmlaffiliation{to}{Department of Computer Science, University of Toronto, Canada}
\icmlaffiliation{co}{Cohere}
\icmlaffiliation{oxstats}{Department of Statistics, University of Oxford, United Kingdom}

\icmlcorrespondingauthor{S\"{o}ren Mindermann}{soeren.mindermann@gmail.com}

\icmlkeywords{Machine Learning, ICML}

\vskip 0.3in
]



\printAffiliationsAndNotice{\icmlEqualContribution} 

\begin{abstract}

\textcolor{red}{A new conference version of this workshop paper is available at:\\ \url{https://arxiv.org/abs/2206.07137}}

We introduce \textit{
Goldilocks
Selection}, a technique for faster model training which selects a sequence of training points that are ``just right’’. We propose an information-theoretic acquisition function---the reducible validation loss---and compute it with a small proxy model---GoldiProx---to efficiently choose training points that 
maximize information about the labels of a validation set.
We show that the ``hard’’ (e.g. high loss) points usually selected in the optimization literature are typically noisy, while the ``easy’’ (e.g. low noise) samples often prioritized for curriculum learning confer less information. Further, points with uncertain labels, typically targeted by active learning, tend to be less relevant to the task. In contrast, \textit{GoldiProx Selection} chooses points that are ``just right’’ and empirically outperforms the above approaches. Moreover, the selected sequence can transfer to other architectures; practitioners can share and reuse it without the need to recreate it.

\end{abstract}

\section{Introduction}
\label{sec:intro}
Models such as GPT-3 \cite{brown2020language}, CLIP \cite{radford2021learning}, and ViT \cite{dosovitskiy2021image} leverage vast quantities of parameters and data to produce remarkable results. Yet, with great size come great computation costs and  prohibitive training times. Accounting further for hyperparameter selection and architecture choice, compute is the core bottleneck that limits performance. Improving computational efficiency paves the way for larger, better performing models. 



Practitioners have long realized that not all samples in the largest datasets are equally useful. Many samples are \textit{noisy}, i.e. mislabelled or inherently ambiguous, so that their label is unpredictable. For example, the next word in a sentence often cannot be foreseen and the caption of a web scraped image is rarely an accurate description. Other samples are \textit{redundant}, for example due to the typical overrepresentation of certain object classes in web scraped data \cite{tian2021divide}. These samples are quickly learned and most could then be ignored without losing performance. Fortunately, data is often so abundant that we can hardly finish a single epoch with the available compute \cite{brown2020language, kaplan2020scaling}, meaning that we can easily afford to skip some of the points.


In the optimization literature, online batch selection methods \cite{loshchilov2015online} train only on points that are `hard' to the current model (e.g. high loss or high gradient norm). 
These techniques include both unbiased importance sampling methods \cite{katharopoulos2018not}, including prioritized experience replay in reinforcement learning \cite{schaul2015prioritized}, as well as biased methods \cite{kawaguchi2020ordered} which simply train on the top-$k$ points with the highest loss or gradient norm. Although online selection can be costly since it computes a forward pass on every point considered, it successfully skips redundant points.


First, we demonstrate that prioritising hard, high-loss examples fails on noisy data. In real-world data, high-loss examples may be mislabelled or inherently ambiguous. We find that, with as little as $10\%$ uniform label noise, high loss points are overwhelmingly those where the label noise is realized. Prioritising them degrades performance severely.

Instead of prioritizing hard points, other bodies of literature including curriculum learning prioritize ``easy'' points with low noise before training on all points equally \cite{bengio2009curriculum}. While this can improve training and convergence, it fails to prioritize points that are not redundant.

A third body of work, active learning, often selects points whose labels are uncertain to the model \cite{houlsby2011bayesian,gal2017deep}. This approach can choose informative points while also avoiding noisy ones. However, we show that it has another weakness: selecting the \textit{less relevant} points that are unlikely to appear at test time, such as those with a low input density $p(x)$. These are unfamiliar to the model and can therefore be selected by active learning approaches. Favouring such obscure points is problematic since they are abundant in uncurated datasets.

To overcome these limitations, we introduce \emph{\textit{Goldilocks Selection}}. Building on recent progress in active learning (anonymous), we introduce an information-theoretic acquisition function---pointwise predictive information gain---which selects points that are ``just right''. By maximizing the information gained about the labels of the validation set, our method chooses non-redundant points without preferring noisy or irrelevant points. Furthermore, we show its equivalence to an intuitive and easy-to-implement novel acquisition function: reducible held-out cross-entropy loss.

Additionally, by using a small proxy model, \emph{GoldiProx Selection} performs online batch selection efficiently. We perform the forward pass for batch selection using a smaller, less compute-intensive model, thereby reducing the computational overhead. Nevertheless, since the proxy sees the same data in the same order as the primary learner, its information state matches the primary model. As such, we find the two models select similar data.

Furthermore, we find that the sequence of training data selected by one small proxy---the GoldiProx Sequence---can accelerate training for models of different size \textit{and} architecture. As such, practitioners can select a sequence one time and reuse it for many training runs, or even download a data sequence created by someone else.



\section{Background: Online batch selection}
\label{sec:background}

Consider training a model $\pof{y\given x,\theta}$ on data $\mathcal{D} = \{(x_i, y_i)\}_{i=1}^n$ using stochastic gradient descent. At each training step $t$, we load a batch $b_t$ of size $|b|$ from $\mathcal{D}$. The labels $\{y_i\}_{i=1}^n$ may be given, as in supervised learning, or could be self-supervised, e.g. the sequence of next words in a document or the rotation an image patch has undergone.

In online batch selection, we pre-sample a larger batch $B_t$ of size $|B|>|b|$ and rank the points $x_i$ in this batch according to a label-aware acquisition function $A(x_i,y_i)$. Then, we construct a smaller batch $b_t$ that consists of the top-ranking $|b|$ points in $B_t$ and perform one gradient step to minimize a mini-batch loss $\sum_{i\in b_t}L(y_i, \pof{y_i\given x_i,\theta})$. The following large batch $B_{t+1}$ is then pre-sampled from $\mathcal{D}$ without replacement of the previously sampled points (all points are replaced at the start of the next epoch).

We use this approach for its simplicity and strong performance in recent work  \cite{kawaguchi2020ordered}. However, it can be extended to incorporate stochastic data selection with importance weights \cite{schaul2015prioritized,katharopoulos2018not} or to include earlier points $x_i$ in $B_t$ whose acquisition score $A(x_i,y_i)$ has been previously computed \cite{loshchilov2015online}. Without importance weights, the simple approach above biases the location of the minimum of the loss. However, this biased selection can improve test performance both in theory and practice \cite{farquhar2021statistical, kawaguchi2020ordered}.

For the full paper see \citet{mindermann2022prioritized}.

\section{What makes points informative?}
\label{sec:acquisition_fn}

\subsection{High loss comes with high noise}

Online batch selection methods typically recommend to train on points with high loss or high gradient norm. Here, we focus on loss because it is significantly easier to compute and correlates with the gradient norm \cite{katharopoulos2018not}. Intuitively, high-loss points can be thought of as `hard' to the current model and therefore not currently redundant (intuitive concepts in this section will be formalized later).

\begin{figure*}[h]
    \centering
    \subfigure[QMNIST]{\includegraphics[width=0.29\textwidth]{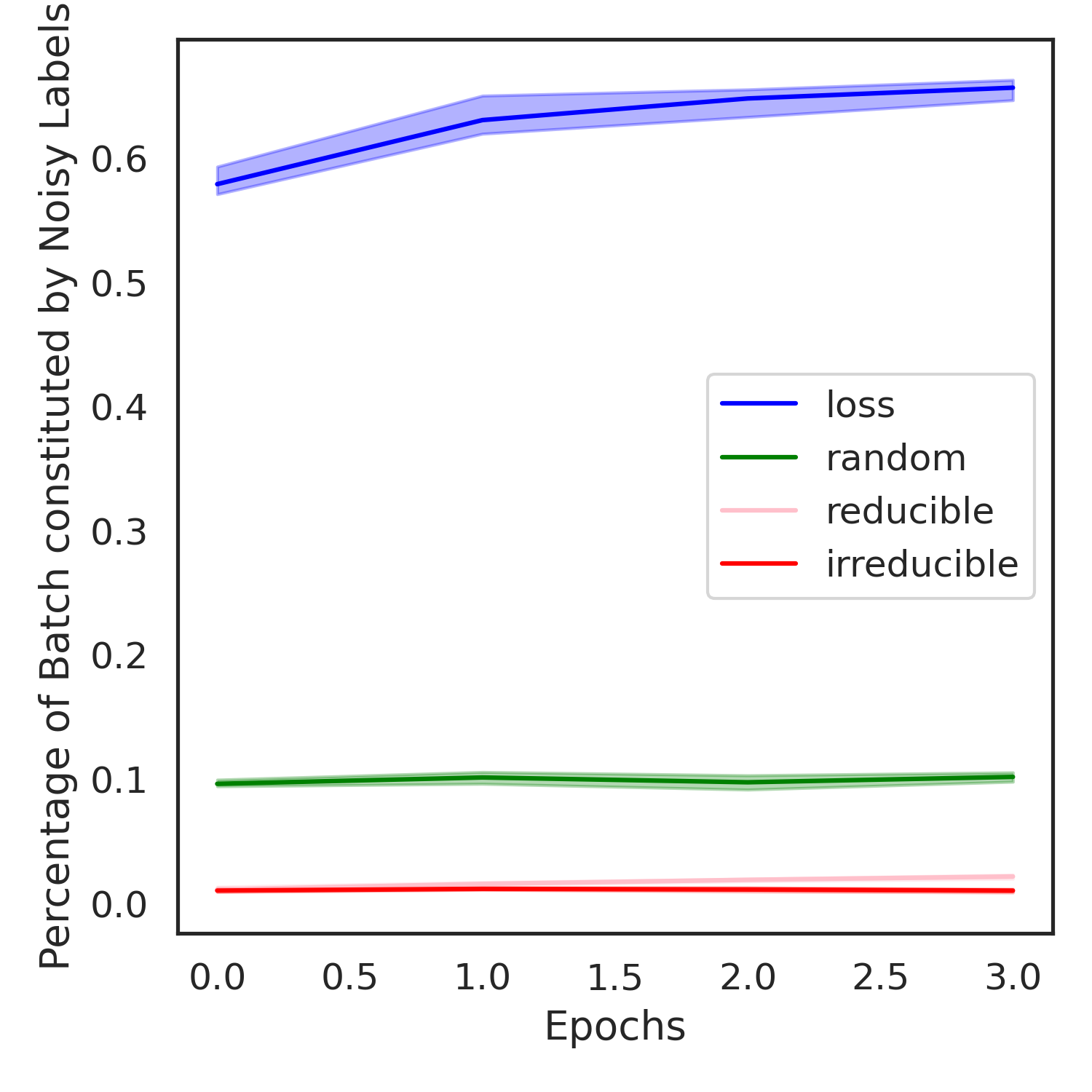}} 
    \subfigure[CIFAR10]{\includegraphics[width=0.285\textwidth]{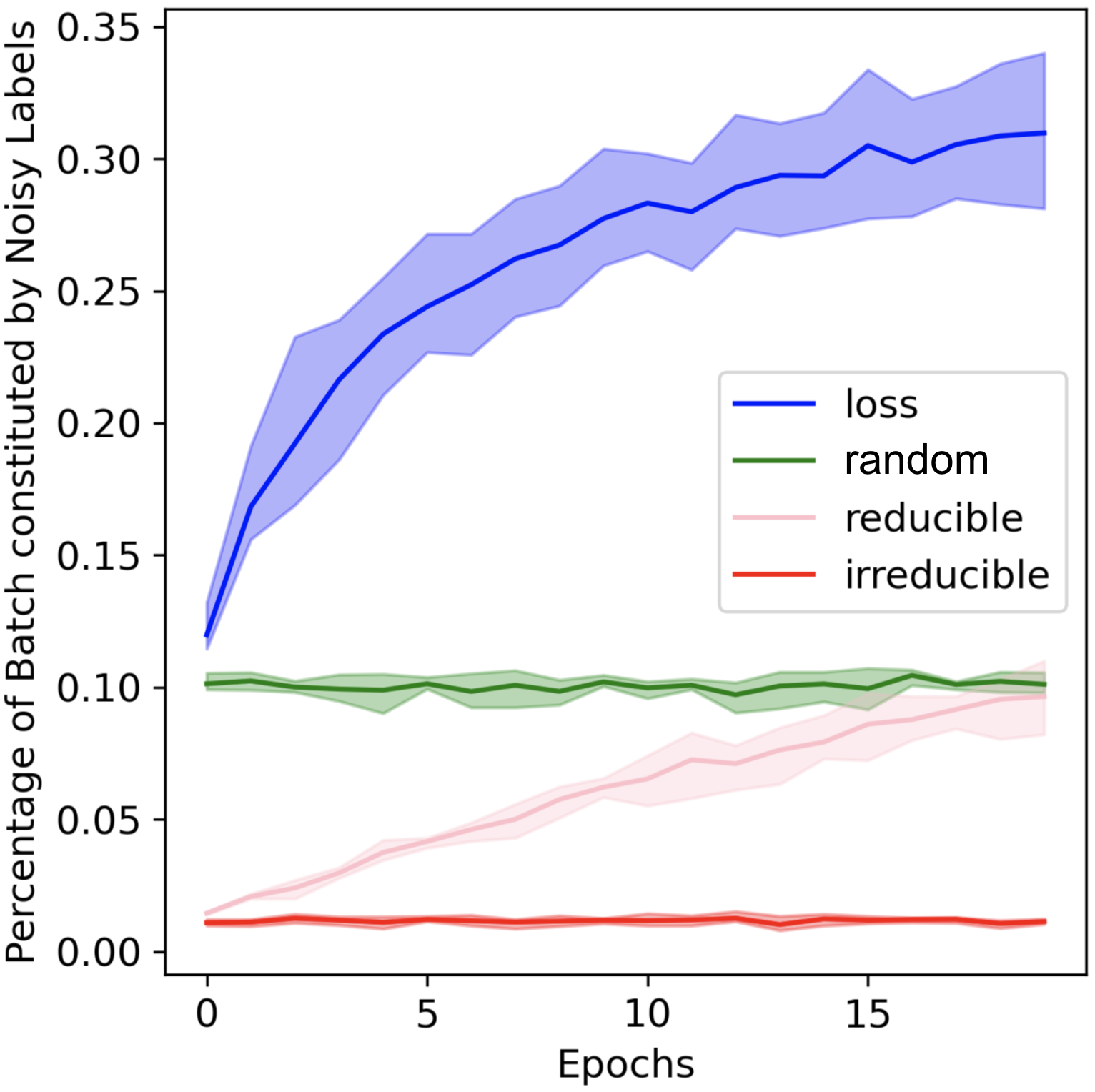}} 
    \subfigure[CINIC10]{\includegraphics[width=0.285\textwidth]{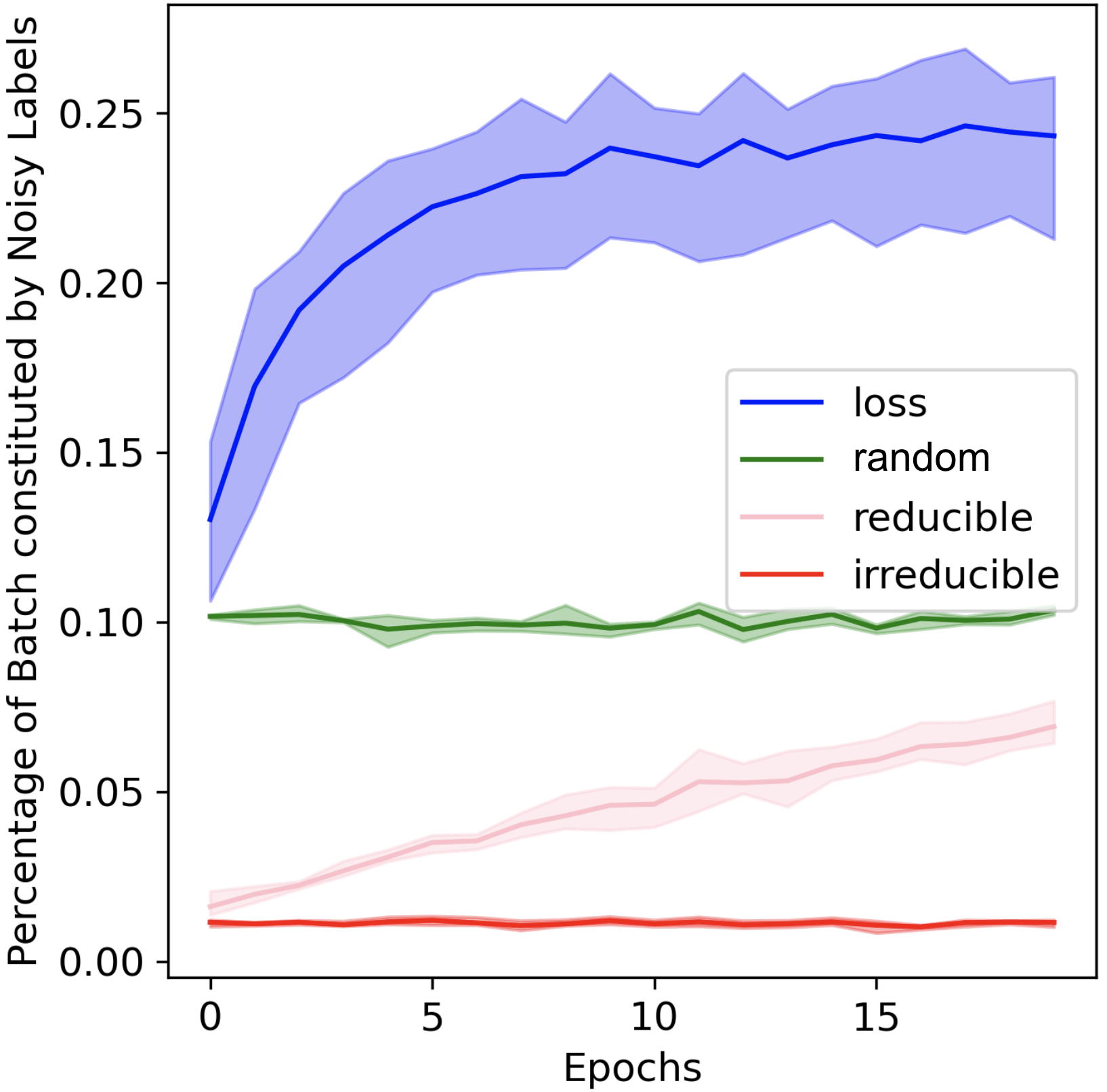}}
    \caption{Percentage of selected points that are corrupted with label noise using different online batch selection criteria. Training data are corrupted with 10\% uniform label noise. Mean along with minimum and maximum across 3 runs shown. Using LeNet model; low epoch setting; no data augmentation.}
    \label{fig:noisy_percentage}
\end{figure*}

\begin{figure*}[h]
    \centering
    \subfigure[QMNIST]{\includegraphics[width=0.293\textwidth]{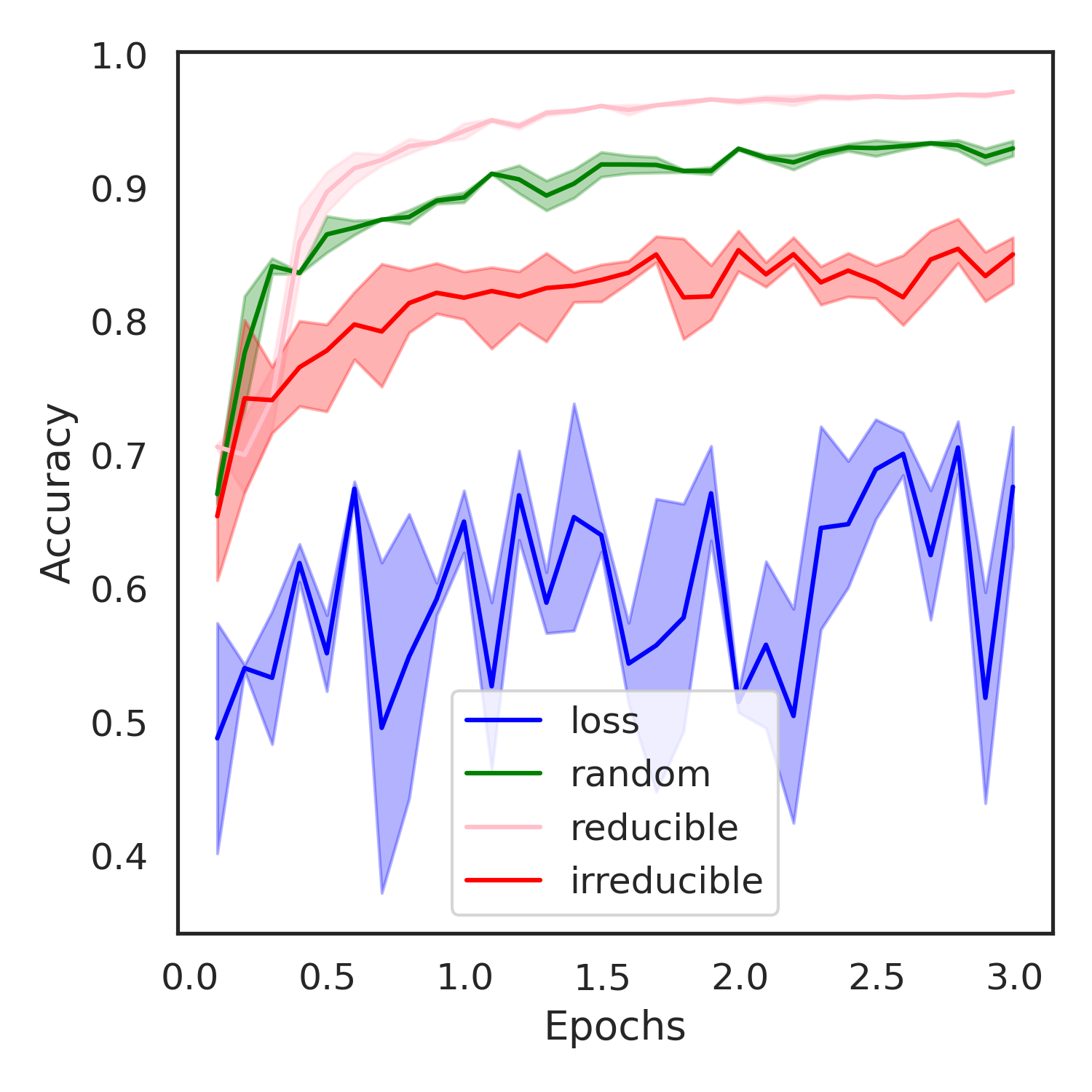}} 
    \subfigure[CIFAR10]{\includegraphics[width=0.285\textwidth]{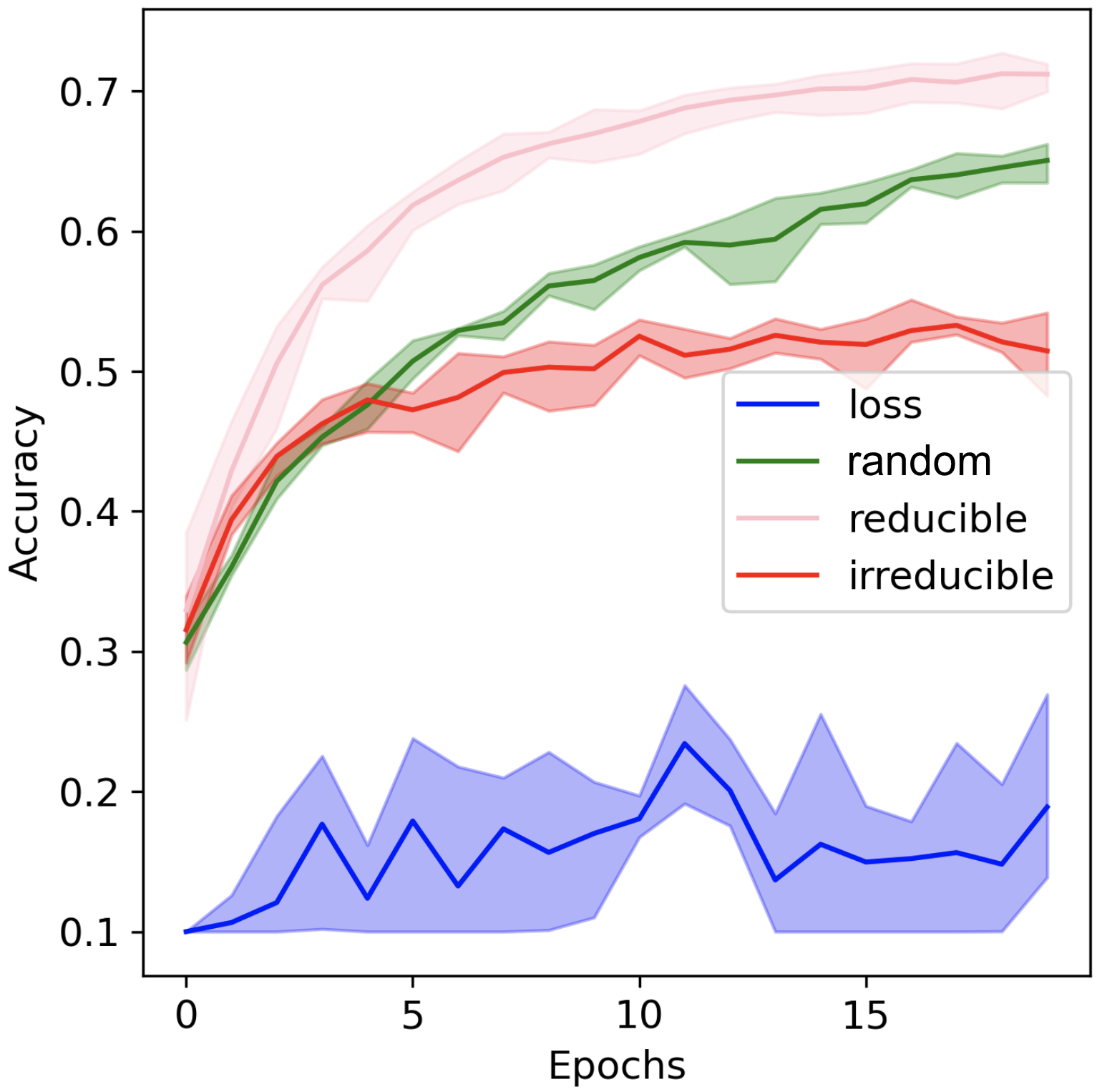}} 
    \subfigure[CINIC10]{\includegraphics[width=0.285\textwidth]{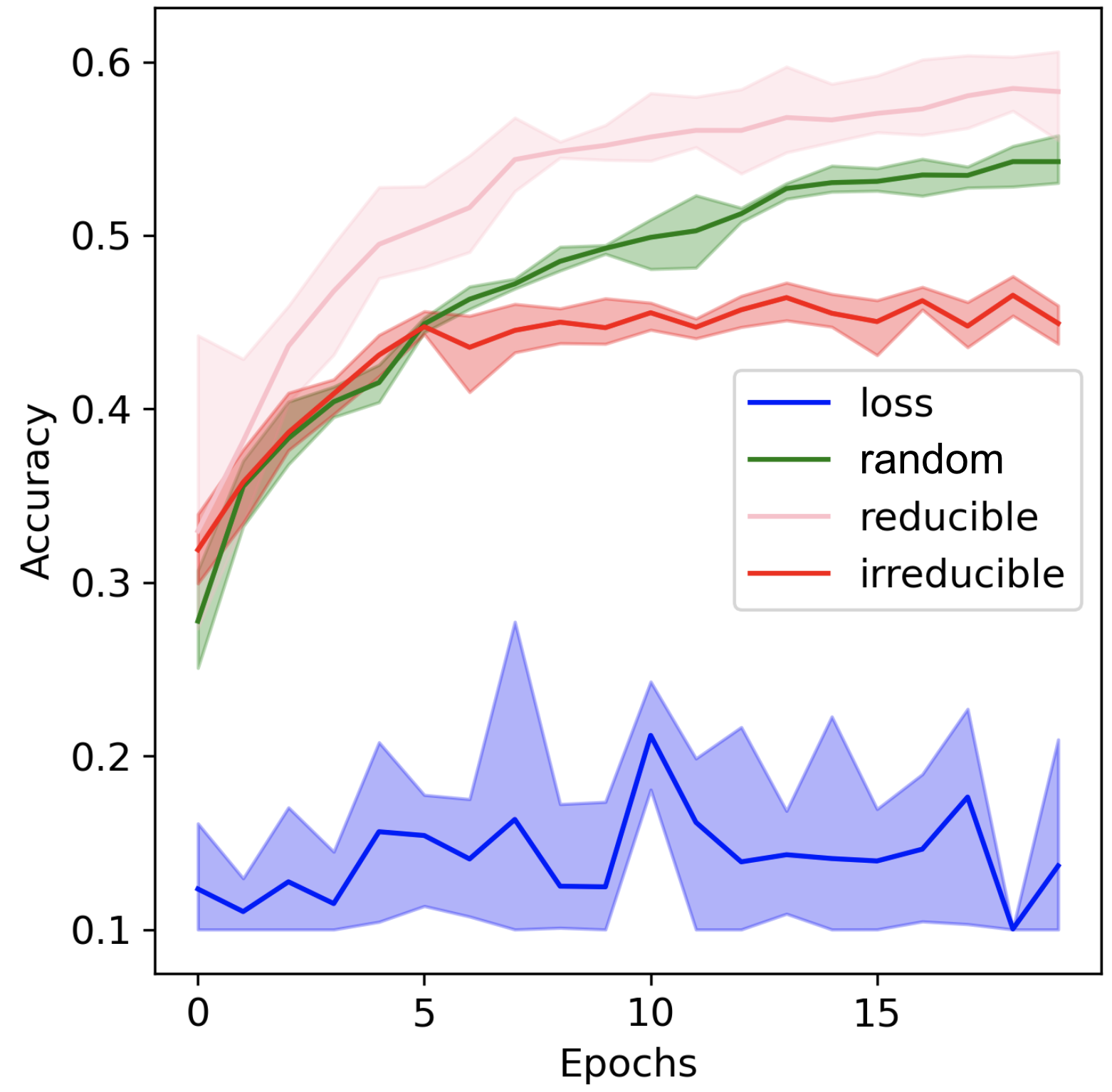}}
    \caption{Image classification test set accuracy for training data with $10\%$ label noise using using different online batch selection criteria. Mean along with minimum and maximum across 3 runs shown. Using LeNet model; low epoch setting; no data augmentation.}
    \label{fig:noisy_accuracy}
\end{figure*}

While loss-based selection can perform well on curated datasets \cite{kawaguchi2020ordered}, we show that it underperforms when adding just $10\%$ uniform label noise\footnote{We can think of a \textit{test time} label as noisy whenever it cannot be predicted with the chosen model class even with unlimited training data \cite{der2009aleatory}. With this definition, noise is common even when labelling is deterministic.}. In Figure \ref{fig:noisy_percentage}, we find that high-loss points (blue line) are overwhelmingly the points mislabelled by the noise distribution across three datasets: QMNIST (same training set as MNIST with a larger test set) \citep{lecun-98, qmnist-2019}, CIFAR-10 \cite{cifar10}, and CINIC-10 (a subset of ImageNet 4.5x larger than CIFAR-10 \cite{darlow2018cinic}). Here, training on the $|b|$ highest-loss points (Figure \ref{fig:noisy_accuracy}, blue) in each large batch $B_t$ degrades performance severely compared to e.g. uniform randomly sampled data (green) and reducible loss (introduced in Section~\ref{sec:theory}).

\subsection{Only removing noisy points is not enough} \label{sec:easy}

Avoiding noisy points is recommended, amongst other things, in the literature on curriculum learning \cite{bengio2009curriculum}. These techniques initially prioritize non-noisy points as noisy points are considered too hard. Eventually, they train on all points in $\mathcal{D}$ equally, although other methods avoid noisy points altogether \cite{thulasidasan2019combating}. 

To probe the effect of prioritizing non-noisy points, we train a separate model $\pof{y\given x, \theta_{\text{val}}}$ to convergence on a validation set $\mathcal{D}_{\text{val}}$ and then evaluate its loss on every point in $\D$. Since a noisy point $x$ likely has high loss even at the end of training (if $x$ was held out during training), we can effectively remove it by using the acquisition function $A(x,y)= - L(y, \pof{y\given x, \theta_{\text{val}}})$  (Figure~\ref{fig:noisy_percentage}, red). However, this does not improve test accuracy (irreducible loss in Figure \ref{fig:noisy_accuracy}, red) compared to random sampling (green) as it has no mechanism to avoid redundant (too easy) points. 

\subsection{Prioritizing information that is relevant to the prediction task}

Active learning typically prioritizes points with uncertain labels\footnote{Instead of label uncertainty, some methods select for data diversity \cite{brinker2003incorporating}. Both criteria help avoid redundancy.}---either measured by a high-entropy model output or by disagreement between models \cite{gal2017deep}. No access to labels is assumed. A prominent approach, Bayesian Active Learning with Disagreement (BALD), even selects informative points that are also predicted to be less noisy \cite{houlsby2011bayesian,gal2017deep}. In a Bayesian model with parameters given by the random variable $\Theta$, it quantifies the expected information (or mutual information) gained about $\Theta$ when observing the unknown label $Y$ for $x$: $I[\Theta; Y | \D_t, x]$.

While observing uncertain points reduces uncertainty about the parameters, we show that it can fail to reduce uncertainty that is relevant to the prediction task. We add 20\% low-relevance points to the data which have uniform white noise input $x$, meaning they have low input density $p(x)$.  Since a low-density point is unlikely to appear at test time, it is not necessarily worth learning.

Indeed, BALD selects these points more than our method. After a warmup period needed to obtain reasonable uncertainty estimates \cite{gal2016dropout}, the white noise samples constitute approximately $20\%$ of the batches selected by BALD (matching the baseline of uniform random sampling).

BALD's behavior can be explained when we consider that more information can be gained about the parameters in areas of the input space where data is still scarce. However, points in such areas have low input density and are therefore unlikely to appear at test time. BALD's tendency to select low-density points is likely to be more extreme in high-kurtosis (heavy tailed) input distributions where a significant share of points are located in low-density areas.

\begin{figure}[h]
    \centering
    \includegraphics[width=0.6\columnwidth]{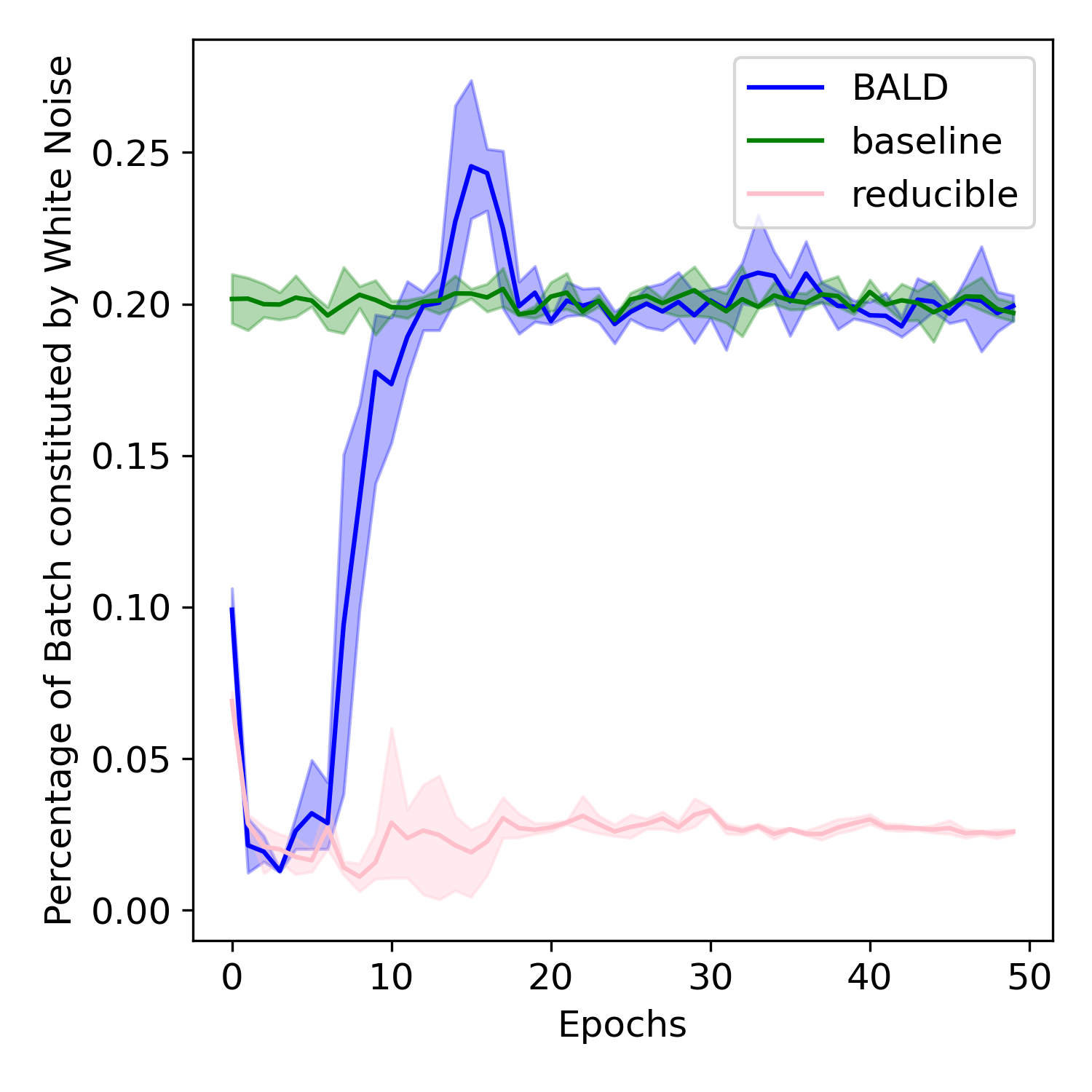}
    \caption{Percentage of selected points on QMNIST that have white noise input (low data density / less likely to appear at test time), using various online batch selection methods.}
    \label{fig:white_noise}
\end{figure}

\section{Experimental details} For experiments on QMNIST, we use a 3-layer MLP with 512 hidden units. For experiments on CIFAR-10 and CINIC-10, we use a small CNN, reminiscent of LeNet \citep{lecun-98}. All models are train using the AdamW optimizer with learning rate $0.001$. In Figure~\ref{fig:white_noise} for BALD (blue) we use $0.005$ learning rate and $0.5$ dropout rate. Hyperparameters are not tuned for the performance of our method. 
Since the goal is to reduce training steps, we evaluate without data augmentation and in the low-epoch setting. Indeed, efficiently training a large model often means training for a single epoch or less without augmentation and stopping far short of convergence \cite{brown2020language,kaplan2020scaling}.

\textbf{Baselines: } We use the \citet{kawaguchi2020ordered} (high loss points) as the external baseline, in addition to random selection and irreducible loss. Some other online batch selection methods \cite{loshchilov2015online, katharopoulos2018not} are complementary to ours and could be combined with our methods in future work. We do not consider curriculum learning, coreset methods, and SVRG \cite{johnson2013accelerating} (these typically cost too much upfront computation when training only for a few epochs, or do not aim to accelerate training) or active learning (which assumes no access to labels).

\section{Pointwise predictive information gain---a measure of reducible loss}
\label{sec:theory}

Based on the previous sections, we are motivated to prioritize points that are learnable\footnote{Can be learned without overfitting to them.} (not noisy) but difficult to the current model and relevant to the validation set. To that end, we propose an information-theoretic acquisition function that maximizes the information gained about the labels of a validation set.

We introduce our method in a Bayesian context. We treat model parameters as a random variable $\Theta$ with prior $\pof{\theta}$ and infer an approximate posterior $\pof{\theta|\D_t}$ using the sequence of already-seen training data $\D_t=b_{1:t-1}$ from the set $\D$. The model has a predictive distribution $\pof{y|x,\D_t} = \int_\theta \pof{y|x,\theta}\pof{\theta|\D_t}d\theta$. However, our method works well with standard non-Bayesian networks which use a point estimate of $\theta$.

The validation set is denoted as $\xval,\yval:=\{x_i\}_{i=1}^{n_{\text{val}}}, \{y_i\}_{i=1}^{n_\text{{val}}}$.

We would like to add a single\footnote{In practice, we add a whole batch $b_t$. Adding multiple points at once is more complex in theory \cite{kirsch2019batchbald} but we can ignore this complication whenever individual training steps cause a negligible change to the model. Acquiring individual points in this case would lead to the same selections.} point $(x,y)$ to $\D_t$ that minimizes the validation cross-entropy loss $\sum_{i\in\D_\text{val}} -\log p(y_i|x_i, \D_t, x, y)$. Thus, we must maximize the information gained about the validation labels.
\citet{kirsch2021evalbald} notes that this information quantity, termed Expected Predictive Information Gain (EPIG), can select informative unlabeled points in the active learning setting. There, the label $y$ for $x$ is not available and so we must compute the expected information gain (also known as mutual information) between $\Yval$ and $Y$:
\begin{align} \label{PIG}
  \MIof{\Yval; Y \given \xval, x, \D_t} &= H[\Yval \given \xval, \D_t] \\
  &- H[\Yval \given \xval, \D_t, x, Y],
\end{align}
which requires computing an expectation over the random variables $Y$ and $\Yval$ (respectively conditioned on $x$ and $\xval$).

Here, we introduce a new quantity, avoiding the need to compute expensive expectations by  making use of the labels $\yval$ and $y$. The label-aware version of the mutual information is known as pointwise mutual information ($\PMIof{\cdot}$); hence we refer to our metric as the pointwise predictive information gain:
\begin{align} \label{PPIG}
  \PMIof{\yval; y \given \xval, x, \D_t} &= \hof{\yval \given \xval, \D_t} \\
  &- \hof{\yval \given \xval, \D_t, x, y}.
\end{align}
The quantities above are the pointwise mutual information and pointwise conditional entropy denoted in lower case:
\begin{align}
    &\hof{y \given x} = -\log \pof{y \given x}; \\
    &\PMIof{y_1,y_2 \given x_1, x_2} := -\log \frac{\pof{y_1 \given x_1} \, \pof{y_2 \given x_2}}{\pof{y_1, y_2 \given x_1, x_2}}.
\end{align}
Note that $\hof{y \given x}$ is just the cross-entropy loss.

This measures the desired quantity: the realized reduction in uncertainty about the validation labels $\yval$ due to observing $(x,y)$. As a result, it satisfies our criteria intuitively: 1) A redundant point (already learned) confers little new information about the validation labels. 2) So does observing a noisy label since a different label may have appeared if the data generation process were repeated. 3) A point that is unlikely to appear in the validation data, and is therefore not relevant to it, also confers little information about it.

\begin{algorithm}[ht!]
	\label{alg:DP-PVI}
	\caption{GoldiProx Selection}
	\newcommand{\StatexIndent}[1][3]{%
  \setlength\@tempdima{\algorithmicindent}%
  \Statex\hskip\dimexpr#1\@tempdima\relax}
	\begin{algorithmic}[1] 
	     \State \textbf{Input:} Initial parameters $\theta^{0}_\text{small}$ and $\theta^{0}_\text{big}$, learning rate $\eta$, small model $\pof{y\given x, \theta_\text{\textbf{val}}}$ trained on validation set, batch size $|b|$, large batch size $|B|>|b|$.
	    \item[]
	    \For{$i$ in \texttt{training set}
	    }
	    \State \texttt{IrreducibleLoss[i]} $\gets L(y_i, \pof{y_i\given x_i, \theta_\text{\textbf{val}}} )$
	    \EndFor
	    \State \texttt{Sequence} $\gets [~]$
	    \item[]
	    \For{$t=1, 2, \ldots$} \Comment{Select data with small model}
	    \State Randomly select a large batch $B_t$ of size $|B|$.
	    \State $\forall i\in B_t$, compute \texttt{Loss[i]}, the loss of point $i$ \NoNumber given parameters $\theta^t_{\text{small}}$ 
	    \State $\forall i\in B_t$, compute \texttt{ReducibleLoss[i]}$\gets$\NoNumber{ $\texttt{Loss[i]}-\texttt{IrreducibleLoss[i]}$} 
	    \State $b_t\gets$ top-$|b|$ samples in $B_t$ in terms of 
	   \NoNumber{$\texttt{ReducibleLoss}$}.
	   \State $g_t\gets$ mini-batch gradient on $b_t$ for $\theta^t_{\text{small}}$
	   \State $\theta^{t+1}_{\text{small}}\gets \theta^t_{\text{small}} -\eta g_t$
	   \State Append $b_t$ to \texttt{Sequence}. 
	    \EndFor
	   \State \For{$t=1, 2, \ldots$} \Comment{Train large model}
	   \State Load $b_t\gets$ \texttt{Sequence[t]}
   	   \State $g_t\gets$ mini-batch gradient on $b_t$ for $\theta^t_{\text{big}}$
	   \State $\theta^{t+1}_{\text{big}}\gets \theta^t_{\text{big}} -\eta g_t$
	   \EndFor
	\end{algorithmic}
\end{algorithm}


However, the two terms in \eqref{PPIG} are still inefficient to compute, as they require training on $(x,y)$ and performing a forward pass on the full validation set. We rewrite \eqref{PPIG} using the symmetry of (pointwise) mutual information, giving:
\begin{align} \label{PPIG2}
 \hof{ y \given x, \D_t} - \hof{y \given x, \xval, \yval, \D_t}.
\end{align}
The first term is simply the cross-entropy loss for $x$ of the current model trained on $D_t$. The second term is the cross-entropy loss of a model trained on $D_t$ and the validation set. Both can be practically computed but we can approximate and further simplify to
\begin{align} \label{PPIG3}
 \hof{ y \given x, \D_t} - \hof{y \given x, \xval, \yval}.
\end{align}
This approximation  is reasonable when $\D_t\ll \D_{\text{val}}$, or when the model $\pof{y\given \yval}$ trained on $\D_{\text{val}}$ has limited capacity to fit the additional data $\D_t$ (see next section).

The second term in \eqref{PPIG3} is simply the acquisition function we introduced in section~\ref{sec:easy}---the  irreducible validation loss. This makes \eqref{PPIG3} an intuitive and theoretically grounded metric: the reducible held-out cross-entropy loss.

\section{Using a small proxy model to select data for other models online}
\label{sec:proxy_and_transfer}
\begin{figure*}[h]
    \centering
    \subfigure[QMNIST]{\includegraphics[width=0.33\textwidth]{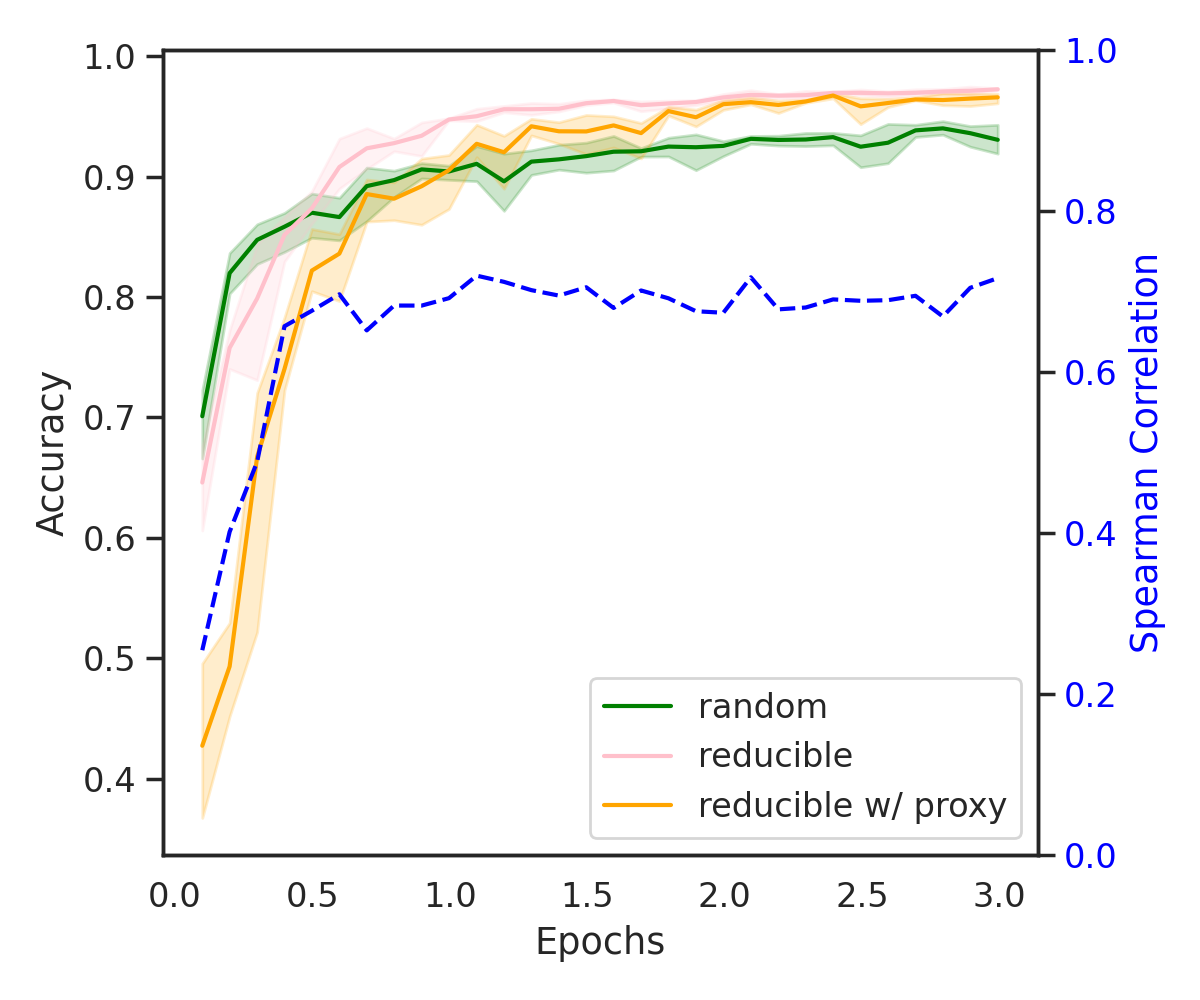}}
    \subfigure[CIFAR10]{\includegraphics[width=0.33\textwidth]{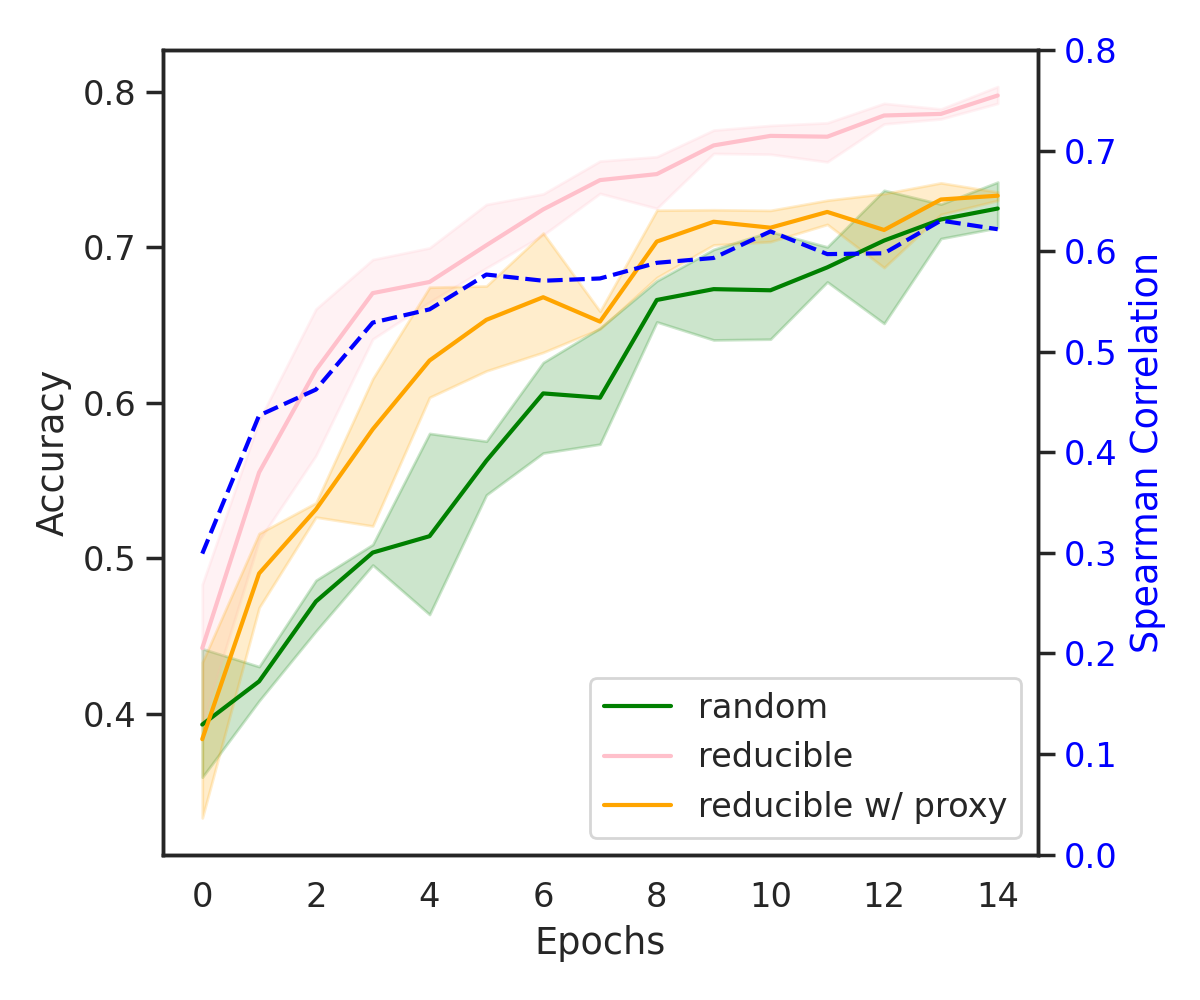}}
    \subfigure[CINIC10]{\includegraphics[width=0.33\textwidth]{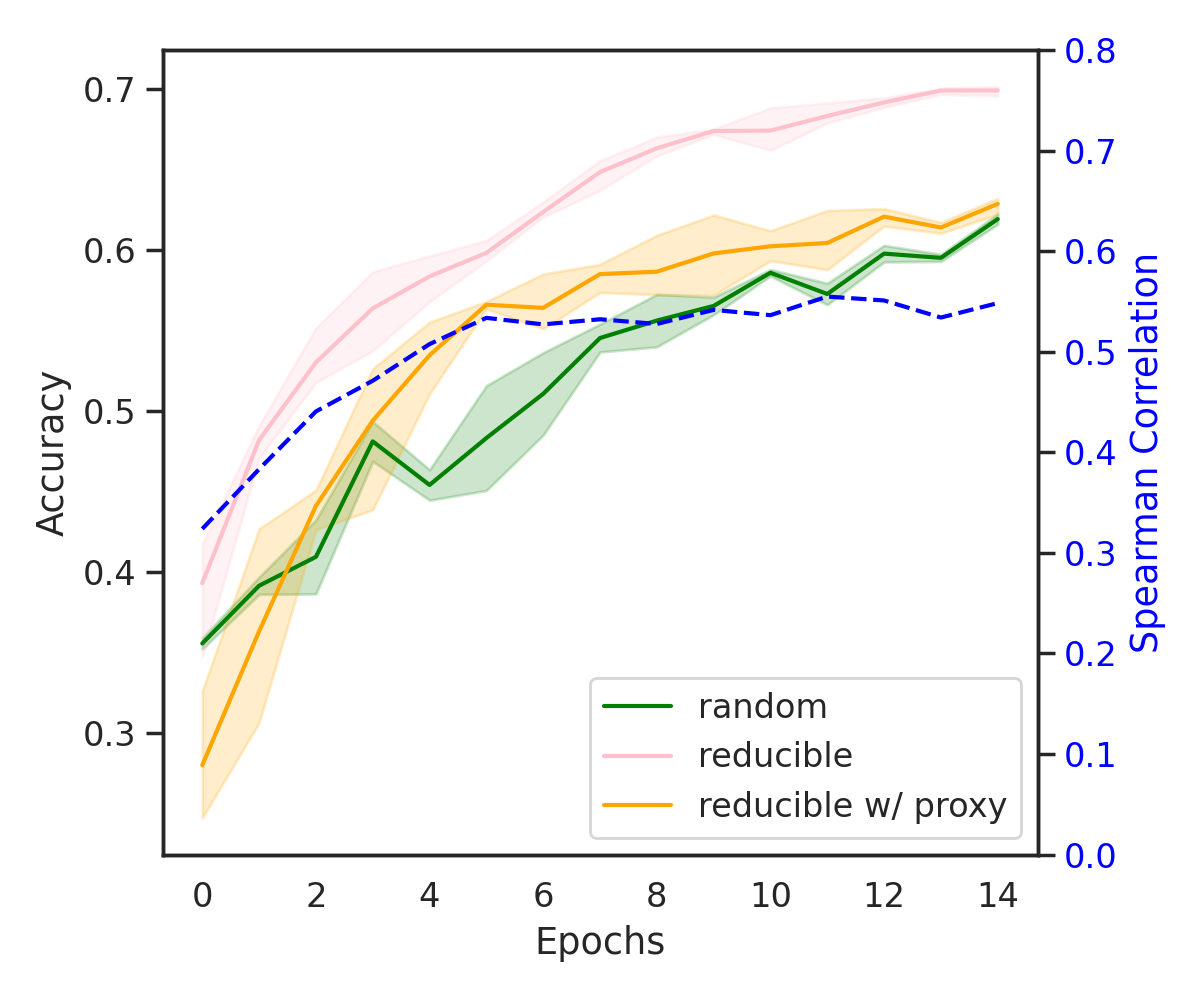}}
        \caption{\textbf{Transfer to larger model with different architecture // rank correlations.} Test set accuracy for image classification tasks using reducible loss with a small proxy and transferring the selected GoldiProx sequence to a large model with different architecture that uses 29x more FLOPs (7x on QMNIST). Methods: Reducible loss (GoldiProx Selection) with and w/o proxy, and random selection. Right axis: rank correlation between reducible loss computed with small proxy model and large model. Positive correlations indicate similar selections.}
    \label{fig:proxy_accuracy}
\end{figure*}
Computational savings from online batch selection have a limit. Compute is only saved by avoiding backward passes. For every point we consider selecting, we need to compute a forward pass to evaluate its acquisition function $A(x_i,y_i)$ \cite{kawaguchi2020ordered, katharopoulos2018not}. Although the forward pass is easy to parallelize and therefore time-efficient, it is not compute-efficient. 

The high cost of data selection has also been studied in the literature on active learning, where \citet{coleman2018select} proposed to select points that are informative as indicated by a fixed small proxy model. This can save drastically more compute, especially when the main model is very large. We build on this approach, extending it to the setting where labels are available and data is selected online rather than with a fixed model.

In the online setting, we must ensure that, at step $t$, the proxy model $\pof{y|x,\theta_{\text{proxy}}}$ selects points informative to the larger model given the large model's information state (described by $\D_t$). Their information states should not differ much. For example, a point that is already learned by a converged model (converged on all of $\D$) may still be informative to a model trained on the sequence $\D_t$ if $x\in \D \setminus \D_t$).


Matching occurs naturally if we store the selected \textit{sequence} of points, not just the set, as shown in Algorithm 1. First, we fully train the proxy using standard online batch selection as described in Section~\ref{sec:background}. We append each selected batch $b_t$ to the data sequence $\D_{t}$ (storing only the indices $i$ of $x_i$ for space efficiency). Subsequently (or simultaneously), we train the large model on $\D_{t}$ in the stored order. 


Matched training on a sequence is simple to implement. For example, one practitioner can create the sequence $\D_{t}$ and another can use it as a plug-in replacement for their existing data loader without changing their training setup.

Empirically, the two matched models do select similar points as indicated by positive rank correlations (Figure~\ref{fig:proxy_accuracy}, blue). Note that perfect correlations are possible even if the larger model outperforms the proxy across all points.

\subsection{Experiments}
We illustrate the effectiveness of our method on QMNIST, CIFAR-10 and CINIC-10.

On QMNIST, we demonstrate that we can use a similar model with small capacity as a proxy model for a large model. We train a 3-layer MLP with 128 hidden units as our small proxy model, and use a 3-layer MLP with 512 hidden units as our large model. This larger MLP has \textbf{6.93$\times$} more parameters than the smaller MLP. Training the large model on the proxy-selected sequence is still significantly faster than the uniform sampling baseline. This can be explained by the high rank correlation of the loss between the proxy model and the large model: the models rank points similarly, providing further validation for the use of a proxy model. As a result, we can save significant computation.

\begin{table}[h]
\centering
\caption{Size and architecture of the proxy and large model used in QMNIST experiments.}
\label{tab:mnist_models}
\begin{tabular}{p{30mm}|ll}
Model                        & Num of Params. & FLOPS   \\ \hline \hline
3-layer MLP with
128 hidden units & 135 k           & 136 k \\
3-layer MLP  with
512 hidden units & 932 k           & 935 k
\end{tabular}
\end{table}

On CIFAR and CINIC, we go one step further by using a proxy model which is not only smaller but also a significantly different and far simpler architecture than the large model. We use a small 1990s-style CNN, reminiscent of LeNet \citep{lecun-98}, as our proxy model, while using ResNet-18 \citep{He2015} as our larger model. ResNet-18 uses approximately \textbf{$29\times$} more floating point operations per second (FLOPS) than the small CNN. This is in addition to using a very different architecture, with residual connections amongst other differences.

\begin{table}[h]
\centering
\caption{Size and architecture of the proxy and large model used in CIFAR10 and CINIC10 experiments.}
\label{tab:cifar_models}
\begin{tabular}{l|ll}
Model & Num of Params. & FLOPS   \\ \hline \hline
Small CNN  & 0.538 M           & 0.019 G \\
ResNet-18  & 11.17 M           & 0.557 G
\end{tabular}
\end{table}

Despite the significant differences, we show that using the sequence obtained using the reducible loss acquisition function with a proxy model, we can speed up training, and thus reduce computational costs. Hence, we can use the selected sequence to train other models. Storing the sequence is cheap since we only save the indices $i$ of each $x_i$. This lends itself well to a variety of settings in which the corpus is large, and computational costs are high which is often the case in low-resource environments. In future work, this should also be extended to large-scale applications such as language modelling and contrastive learning, to reduce their training speed and computation.


\newpage
\clearpage
\bibliography{references}
\bibliographystyle{icml2021}





\end{document}